\documentclass[letterpaper, 12pt]{article}
\usepackage{graphicx}
\usepackage{times}
\usepackage{amsmath, amssymb}
\usepackage{geometry}
\usepackage{titlesec}
\usepackage{caption}
\usepackage{float}
\usepackage{booktabs}
\usepackage{url}
\usepackage{enumitem}
\usepackage[most]{tcolorbox}
\usepackage{float} 
\usepackage{xcolor}

\titleformat{\section}{\large\bfseries}{}{0pt}{}
\titleformat{\subsection}{\normalsize\bfseries\itshape}{}{0pt}{}
\titleformat{\subsubsection}{\normalsize\bfseries}{}{0pt}{}

\begin{document}

\begin{center}
    {\Large \textbf{Text to Automata Diagrams: Comparing TikZ Code Generation with Direct Image Synthesis}} \\
    \vspace{0.5em}
    {\large Ethan Young$^{1}$,  Zichun Wang$^{2}$, Aiden Taylor$^{1}$, Chance Jewell$^{1}$, Julian Myers$^{1}$, Satya Sri Rajiteswari Nimmagadda$^{1}$, Anthony White$^{2}$, Aniruddha Maiti$^{2}$, Ananya Jana$^{1}$}\\
    \vspace{0.3em}
    {$^{1}$Marshall University \quad $^{2}$West Virginia State University}
\end{center}

\begin{abstract}
Diagrams are widely used in teaching computer science courses. They are useful in subjects such as automata and formal languages, data structures, etc. These diagrams, often drawn by students during exams or assignments, vary in structure, layout, and correctness. This study examines whether current vision-language and large language models can process such diagrams and produce accurate textual and digital representations. In this study, scanned student-drawn diagrams are used as input. Then, textual descriptions are generated from these images using a vision-language model. The descriptions are checked and revised by human reviewers to make them accurate. Both the generated and the revised descriptions are then fed to a large language model to generate TikZ code. The resulting diagrams are compiled and then evaluated against the original scanned diagrams. We found descriptions generated directly from images using vision-language models are often incorrect and human correction can substantially improve the quality of vision language model generated descriptions. This research can help computer science education by paving the way for automated grading and feedback and creating more accessible instructional materials.

\end{abstract}

\section{Introduction}

Automata diagrams appear in undergraduate computer science courses that cover formal languages and computation.  Students use hand-drawn sketches to take notes, to solve problems, to explain processes, or to show how systems behave. These diagrams describe the structure of finite automata through labeled states and transitions. Students draw these sketches by hand during exams, quizzes, and assignments. The drawings are often informal and vary in layout, notation, and correctness. Vision language models can process scanned diagrams and generate text. The large language models (LLM) can take structured text and produce LaTeX TikZ code. TikZ is a typesetting language commonly used for diagrams in instructional material. The sequence from diagram to description to TikZ can be used to reconstruct a student's diagram in a clean format. This reconstruction depends on both the accuracy of the generated text and the ability of the model to follow the diagram conventions used in computer science theory subjects such as automata.

This paper studies the reliability of such a reconstruction process. The experiments use a collection of student-drawn diagrams. A description for each image is generated by a vision-language model (GPT-4o). We found that such descriptions are not always accurate. For this reason, human review is performed to revise the descriptions to correct errors in these descriptions. Both types of descriptions are used as input to a large language model to generate corresponding TikZ code. The TikZ output is compiled and visually compared with the original sketch.

The study includes three comparisons. First, we compared the semantic content of the raw and edited descriptions using similarity scores to understand how much they differ. Second, we compared the diagrams generated from raw descriptions and edited descriptions with the student sketches in order to understand if human edits can improve vision-language model generated descriptions. Third, we compared the diagrams generated from TikZ code (LLM generated code) of raw descriptions and from TikZ code (LLM generated code) of edited descriptions with the student sketches. The results show that the diagrams generated either directly from the edited descriptions or from the TikZ code (LLM generated code) of the edited descriptions are more accurate.

\section{Related Work}

Scientific figures play a central role in research dissemination and teaching. Many figures contain dense structure, domain rules, and visual conventions that differ from natural images. Recent work on figure-to-text generation involves constructing new datasets and systems such as SciCap \cite{hsu2021scicap} and SciCap+ \cite{yang2024scicap+}. Similarly, there are works on text-to-figure generation such as TikZero \cite{belouadi2025tikzero}, TikZilla \cite{learningtikzilla}, ScImage \cite{zhang2024scimage}, and Pluto \cite{srinivasan2025pluto}.  Recent work has introduced benchmarks such as AnaFig \cite{yue2025anafig}, Chart-HQA \cite{chen2025chart}, ChartCoder \cite{zhao2025chartcoder}, and MatCha \cite{lai2025can}.  There are related studies which include evaluation methods such as DiagramEval \cite{liang2025diagrameval} and the Visual Consistency Score from ChartCap \cite{lim2025chartcap}. Large datasets such as ChartCap \cite{lim2025chartcap}, PatentDesc-355K \cite{shukla2025patentlmm}, Chart2Code-160k \cite{zhao2025chartcoder}, and DaTikZ-V4 \cite{learningtikzilla} have been released to support diagram understanding by artificial intelligence models.  
Several studies were performed in different domains such as patents, materials science, and nuclear imagery \cite{shukla2025patentlmm,joynt2024comparative,lai2025can}. Two survey papers explored the broader domain of image and chart \cite{sordo2025review,yan2025chart}.  A majority of these works focus on generic or domain specific diagrams. In this work, we focus on diagrams drawn specifically for the automata course, as such domain specific diagrams offer a rich set of examples involving multiple levels of logic

\section{Dataset Preparation}
\subsection{Image scan collection}

The image data used in this study consists of scanned diagrams drawn by students in undergraduate automata theory courses. The diagrams were collected from exams and assignments from those course offerings. Each image corresponds to a student response to a specific problem that required drawing a finite automaton or a related state-based diagram. Since multiple students answered the same question, there are several distinct diagrams for a single problem. The collection includes both correct and incorrect student responses. We did not attempt to filter diagrams based on correctness at the time of collection. The scanned images were stored in standard image formats and later consolidated into a unified corpus for analysis. The final corpus contains approximately 190 diagrams spanning 28 distinct exam and assignment questions. The diagrams include deterministic finite automata, nondeterministic finite automata with $\varepsilon$-transitions and machines, Pushdown automata, graphs, trees, and Turing machines. In these diagrams, layout, notation, and level of detail significantly vary.

The use of student-submitted material in this study was reviewed and approved under an Institutional Review Board (IRB) at Marshall University. All images were anonymized prior to analysis. No identifiable student information was retained, stored, or used at any stage of the study.

\subsection{Prompt Design for Description Generation}

Three types of prompts were tested for generating text descriptions from student-drawn diagrams.

\begin{itemize}[noitemsep]
    \item \textbf{Diagram-only prompt}: The model received only the diagram image. No additional context was given.
    
    \item \textbf{Question-conditioned prompt}: The model received the image and the original exam question. This was intended to reduce ambiguity in cases where the diagram alone was insufficient.

    \item \textbf{One-shot prompt}: The model received the image, a similar example diagram, and a corresponding example caption.
\end{itemize}

The following is an example of a question-conditioned prompt:

\begin{tcolorbox}[
  colback=gray!10,
  colframe=gray!50!black,
  boxrule=0.5pt,
  arc=3pt,
  boxsep=5pt,
  left=6pt,
  right=6pt,
  top=6pt,
  bottom=6pt,
  enhanced,
  width=\linewidth,
  halign=justify
]
You are given a scanned image of a student-drawn diagram. The exam question is: “Draw a finite automaton that accepts all strings over $\{0,1\}$ containing an even number of 0s and an even number of 1s.” Describe the diagram in plain English. Your description should include all states, transitions, and labels exactly as they appear. If any detail is unclear, state this explicitly. Do not explain or interpret the diagram. Do not add content that is not visible.
\end{tcolorbox}

We found that the question-conditioned prompts led to more consistent structure in the generated text. However, missing transitions and incorrect state roles still appeared in some outputs. The one-shot prompt performed better when the example closely matched the target question.

\subsection{Generation of Human Edited Version of the Description}

Each model-generated description was reviewed by a human annotator. Edits focused on correcting structural inaccuracies, clarifying ambiguous references, and aligning the description with the visual content. Both the raw and edited versions were retained to evaluate the effect of revision on downstream TikZ code generation.

\begin{figure}[H]
    \centering
    \includegraphics[width=0.6\linewidth]{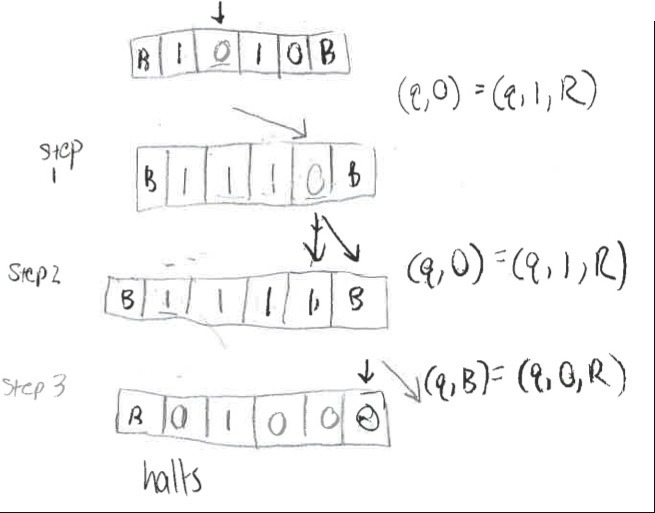}
    \caption{Sample of student-drawn diagram related to Turing machine execution trace}
    \label{fig:im0464}
    
    \vspace{1em}
    
    \begin{tcolorbox}[
      colback=gray!5,
      colframe=black!50,
      title=Vision Language Model Generated Description,
      fonttitle=\bfseries,
      boxrule=0.5pt,
      arc=3pt,
      boxsep=4pt,
      left=6pt,
      right=6pt,
      top=6pt,
      bottom=6pt,
      enhanced,
      width=\textwidth,
      halign=justify
    ]
    \small
    A sequence of three horizontal rows of boxes is shown, each containing five boxes. The first row has boxes labeled from left to right as B, 1, 0, 1, B, with a downward arrow above the 1. To the right of this row, there is a label "(q,0) = (q,1,R)". The second row has boxes labeled B, 1, 1, 0, B, with a downward arrow above the 1 and another downward arrow between the 1 and 0. To the right of this row, there is a label "(q,0) = (q,1,R)". The third row has boxes labeled B, 0, 1, 0, B, with a downward arrow above the B and a circular symbol with a diagonal line through it to the right of the last box. To the right of this row, there is a label "(q,B) = (q,0,R)". The word "halts" is written below the third row. Each row is connected by straight downward arrows from the first to the second and from the second to the third. The rows are labeled on the left as step 1, step 2, and step 3, respectively.
    \end{tcolorbox}
    \begin{tcolorbox}[
      colback=blue!5,
      colframe=blue!50,
      title=Human-Edited Description,
      fonttitle=\bfseries,
      boxrule=0.5pt,
      arc=3pt,
      boxsep=4pt,
      left=6pt,
      right=6pt,
      top=6pt,
      bottom=6pt,
      enhanced,
      width=\textwidth,
      halign=justify
    ]
    \small
    A sequence of four horizontal rows of boxes is shown, each containing six boxes. The first row has boxes labeled from left to right as B, 1, 0, 1, 0, B, with a downward arrow above the first 0. The first 0 is also underlined. To the right of this row, there is a label "(q,0) = (q,1,R)". The second row has boxes labeled B, 1, 1, 1, 0, B, with a right and downward arrow from below the upper row’s first 0 to the second row’s first 0. This 0 in the second row is underlined. To the right of this row, between the second row and the third row horizontally, there is a label "(q,0) = (q,1,R)". The third row has boxes labeled B, 1, 1, 1, 1, B, with a downward arrow above the last 1 and above the last B. To the right of this row, between the third row and the fourth row there is a label "(q,B) = (q,0,R)". The word "halts" is written below the fourth row. The fourth row has boxes labeled B, 0, 1, 0, 0, 0. There is a downward pointing arrow above the final 0. There is a diagonal rightward pointing arrow to the right of this downward arrow. Each row is connected by straight downward arrows from the first to the second and from the second to the third. To the left of row two is the text "Step 1", to the left of row three is the text "Step 2", and to the left of row four is the text "Step 3."
    \end{tcolorbox}
\end{figure}

An example illustrating a student-drawn diagram, the corresponding model-generated description, and the human-edited description is shown along with Figure-\ref{fig:im0464}.

\section{Experiments}
In our experiments, GPT-4o is used to perform both tasks : the Vision-Language Model task and the Large Language Model task \cite{openai2024gpt4o}.
\subsection{Semantic Distance Between Descriptions}

The difference between raw and edited descriptions is measured using cosine similarity between sentence-level embeddings. Higher similarity scores correspond to minor edits.  Lower scores indicate that significant edits are performed on the generated description by human reviewer. The cosine similarity metric serves as a proxy for estimating the extent of human intervention required to correct model-generated text.

\subsection{Human Evaluation of Images Generated Using Descriptions}
\label{sucsec:human_eval_text}
Since visual comparison between two diagrams is more direct and less ambiguous than comparison between textual descriptions, image based human evaluation was conducted to assess the quality of model generated descriptions before human revision and the corresponding human edited descriptions. Images were generated from both description types using a vision language model. For each original hand-drawn diagram, two generated diagrams were produced: one from the model generated description before revision and one from the human edited description. Human evaluation was conducted to measure agreement between each generated diagram and the corresponding hand-drawn diagram. Two human evaluators independently inspected each pair and assigned scores using a 5-point Likert type scale \cite{sullivan2013analyzing} following the strategy adopted in \cite{yue2025anafig}. 
If the difference between the two evaluators’ scores was greater than one point on the 5-point scale, the two evaluators met and assigned a single agreed final score. The evaluators assigned a score to each generated diagram according to the following rubric:

\begin{itemize}
    \item \textbf{1} - Completely incorrect: the generated diagram represents a different automata machines.
    \item \textbf{2} - Largely inconsistent: most structural elements do not match.
    \item \textbf{3} - Partially aligned: major states are present, but key transitions or accepting states are missing.
    \item \textbf{4} - Mostly consistent: minor structural or labeling errors.
    \item \textbf{5} - Fully consistent: states, transitions, and accepting conditions match.
\end{itemize}

During independent evaluation, the absolute difference between the two evaluators’ scores was at most one point in more than 95\% of the evaluated diagrams.

\subsection{Human Evaluation of TikZ Compiled Diagrams}
We also performed an indirect TikZ based evaluation. Both the model generated descriptions before human revision and the human edited descriptions are submitted to a language model to produce TikZ code. The generated codes are compiled to obtain diagrams. Diagram level human evaluation is then conducted to measure agreement between each compiled diagram and the corresponding hand-drawn image. Two evaluators assigned a score on a 5-point scale using the same rubric as described in the previous subsection.

\section{Results}
\subsection{Semantic Similarity Between Vision-Language Generated Text and Human-Edited Description}

To quantify the difference between raw model-generated descriptions and their human-edited counterparts, we compute standard text similarity metrics including BLEU, METEOR, and ROUGE-L. These metrics capture complementary aspects of overlap between the two versions, such as n-gram precision (BLEU), token-level alignment with synonym handling (METEOR), and longest common subsequence similarity (ROUGE-L).

Scores are computed for each matching pair of raw and edited descriptions and then averaged across the full dataset. Lower scores indicate descriptions that required substantial human revision due to missing states, incorrect transitions, or ambiguous structure, while higher scores correspond to descriptions that required only minor edits such as labeling or formatting corrections.

Table~\ref{tab:avg_semantic_scores} reports the average similarity scores across all caption pairs. The relatively high ROUGE-L and METEOR values indicate that human edits typically preserve much of the original lexical content while correcting structural inaccuracies, whereas the lower BLEU score reflects sensitivity to reordering and phrasing changes common in structural revisions.

\begin{table}[H]
\centering
\begin{tabular}{lccc}
\toprule
Metric & BLEU & METEOR & ROUGE-L \\
\midrule
Average Score & 0.570 & 0.680 & 0.730 \\
\bottomrule
\end{tabular}
\caption{Average semantic similarity scores between raw and edited descriptions.}
\label{tab:avg_semantic_scores}
\end{table}

\subsection{TikZ Compilation Rate}
We were able to compile 84 of the TikZ generated from the human edited descriptions and 98 of the TikZ generated from the original descriptions in the first round. GPT-4o was generating some extra characters although our prompt asked it not to generate \cite{openai2024gpt4o}. We create a program to clean up those extra characters at the beginning and end of the tex file. We found that the generated TikZ code was using some additional packages and including those additional packages increased the compilation rate. We were able to compile 129 of the TikZ code generated from the edited descriptions and 131 of the TikZ code generated from the original descriptions. 

\subsection{Human Evaluation of Reconstruction Quality Using Diagrams Directly Generated from Text Descriptions}

We performed a lightweight evaluation of 20 diagrams (10\% of the dataset) for the direct diagram generation from description. 
\begin{table}[h]
\centering
\begin{tabular}{lcc}
\textbf{Comparison} & \textbf{Avg. Score (out of 5)} & \textbf{Std. Dev.} \\
\hline
Diagram from Edited Description vs. Hand-drawn & 3.6 & 1.09 \\
Diagram from Original Description vs. Hand-drawn & 2.85 & 1.3 \\
\end{tabular}
\caption{Human evaluation scores comparing generated diagrams with hand-drawn diagrams.}
\label{tab:human_scores}
\end{table}
We compared the diagrams generated from the edited description with the hand-drawn diagram and the diagrams generated from the original description with the hand-drawn diagram. The results are shown in the Table~\ref{tab:human_scores}.
The results indicate that the human reviewers assigned higher scores to diagrams produced from edited descriptions. 
Some of the errors in diagrams generated from raw descriptions included:

\begin{itemize}[noitemsep]
    \item Missing or extra transition
    \item Incorrect accepting states
    \item Misplaced labels and loops
\end{itemize}

\subsection{Human Evaluation of Reconstruction Quality Using Diagrams Generated Via TikZ Code}
Since full dataset evaluation would require manual visual inspection of each compiled diagram against the corresponding hand drawn diagram, which is time consuming, requires substantial manual effort and incurs additional API costs for repeated vision language model calls, a 10 percent subset (20 diagrams) was selected for scoring. We compared the TikZ compiled diagrams  from the edited description with the hand-drawn diagram and the TikZ compiled diagrams  from the original description with the hand-drawn diagram.  Table~\ref{tab:human_scores2}
\begin{table}[h]
\centering
\begin{tabular}{lcc}
\textbf{Comparison} & \textbf{Avg. Score (out of 5)} & \textbf{Std. Dev.} \\
\hline
Tikz Diagram from Edited Description vs. Hand-drawn & 4.65 & 0.48 \\
Tikz Diagram from Original Description vs. Hand-drawn & 2.95 & 1.14 \\
\end{tabular}
\caption{Human evaluation scores comparing TikZ compiled diagrams with hand-drawn diagrams.}
\label{tab:human_scores2}
\end{table}

As in the case of directly generated image comparisons, some of the errors observed in diagrams generated from raw descriptions via TikZ included cases where transitions were missing or additional transitions were introduced, accepting states were incorrectly designated, and transition labels or self loops were placed on incorrect states or edges relative to the corresponding hand drawn diagram.

\subsection{Subjective Evaluation}
\begin{figure}[h]
\centering
\includegraphics[width=0.15\textwidth]{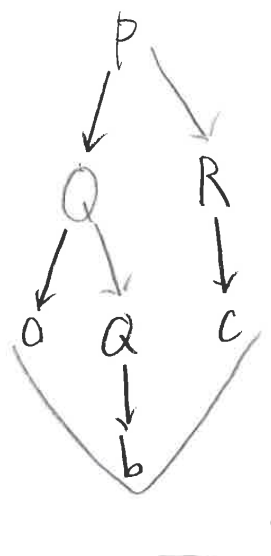}
\hfill
\includegraphics[width=0.3\textwidth]{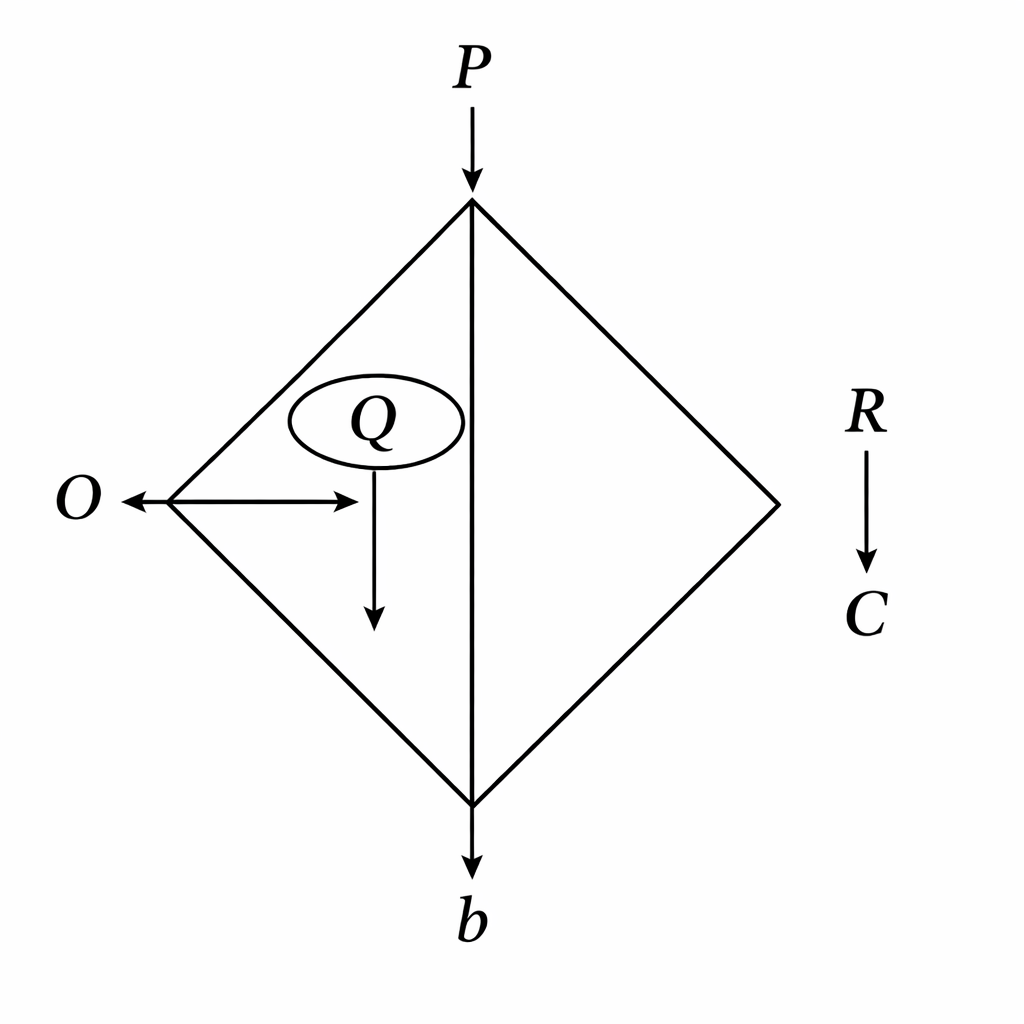}
\hfill
\includegraphics[width=0.3\textwidth]{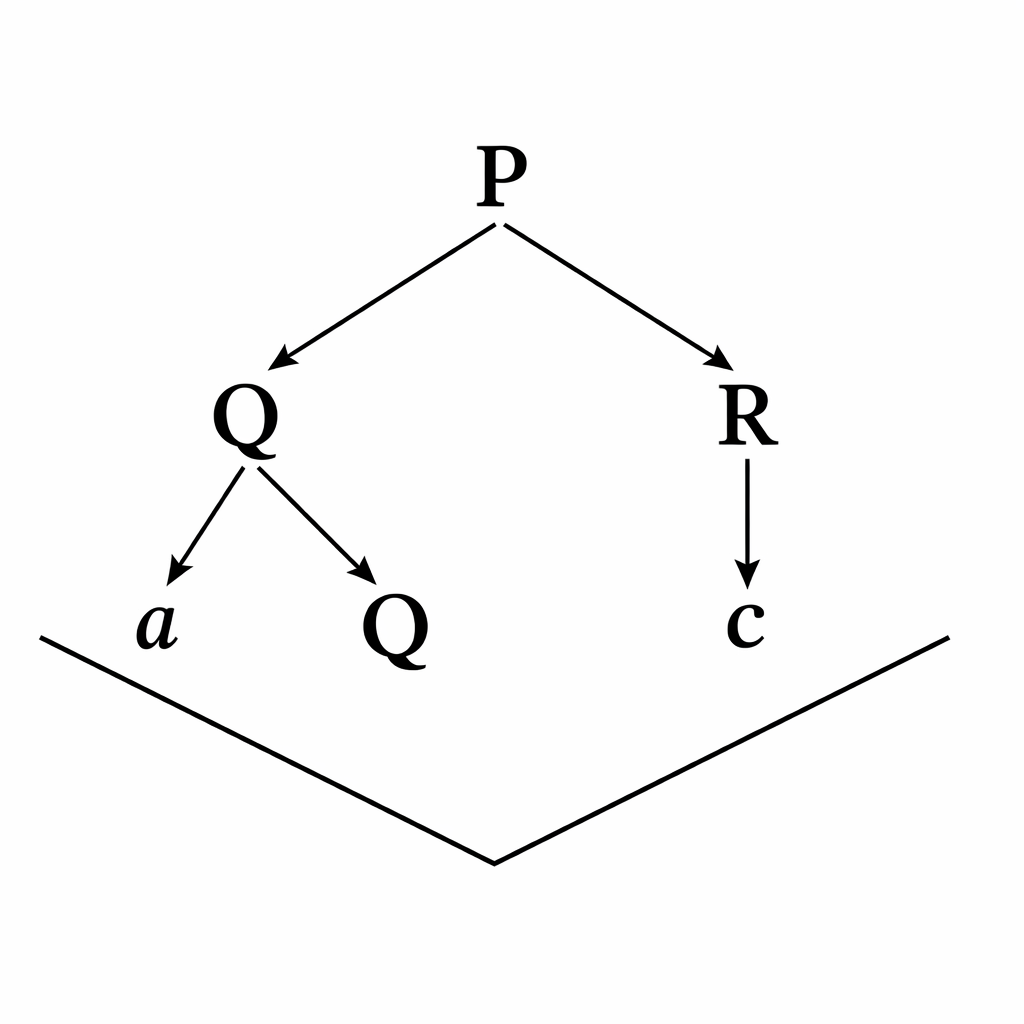}

\caption{Hand-drawn diagram (left), diagram from original description (middle) and diagram from edited description (right)}
\label{fig:dia1}
\end{figure}

\begin{figure}[h]
\centering
\includegraphics[width=0.3\textwidth]{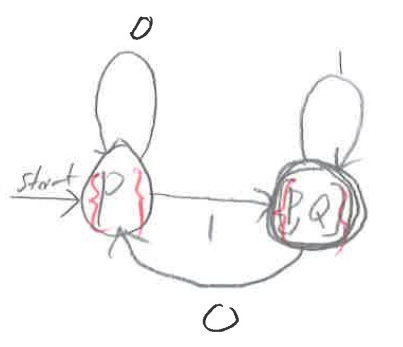}
\hfill
\includegraphics[width=0.25\textwidth]{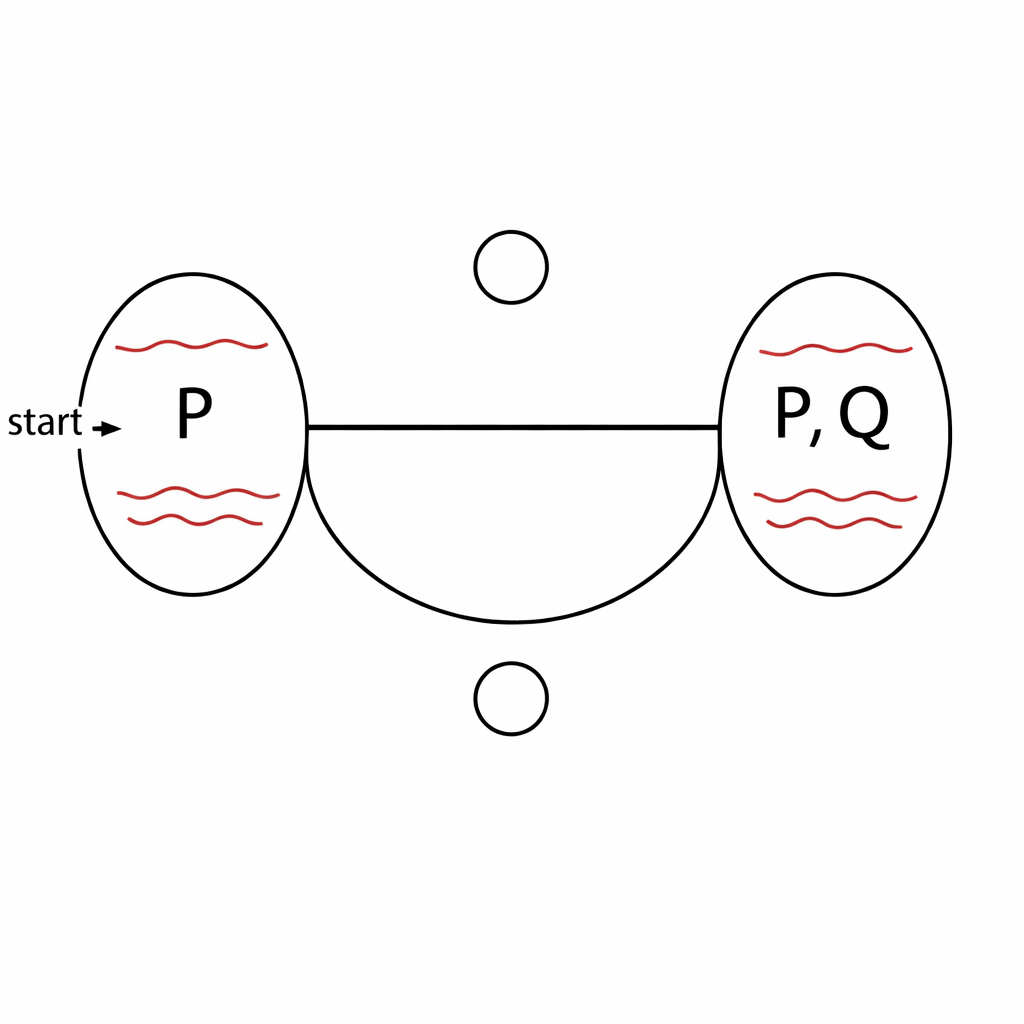}
\hfill
\includegraphics[width=0.35\textwidth]{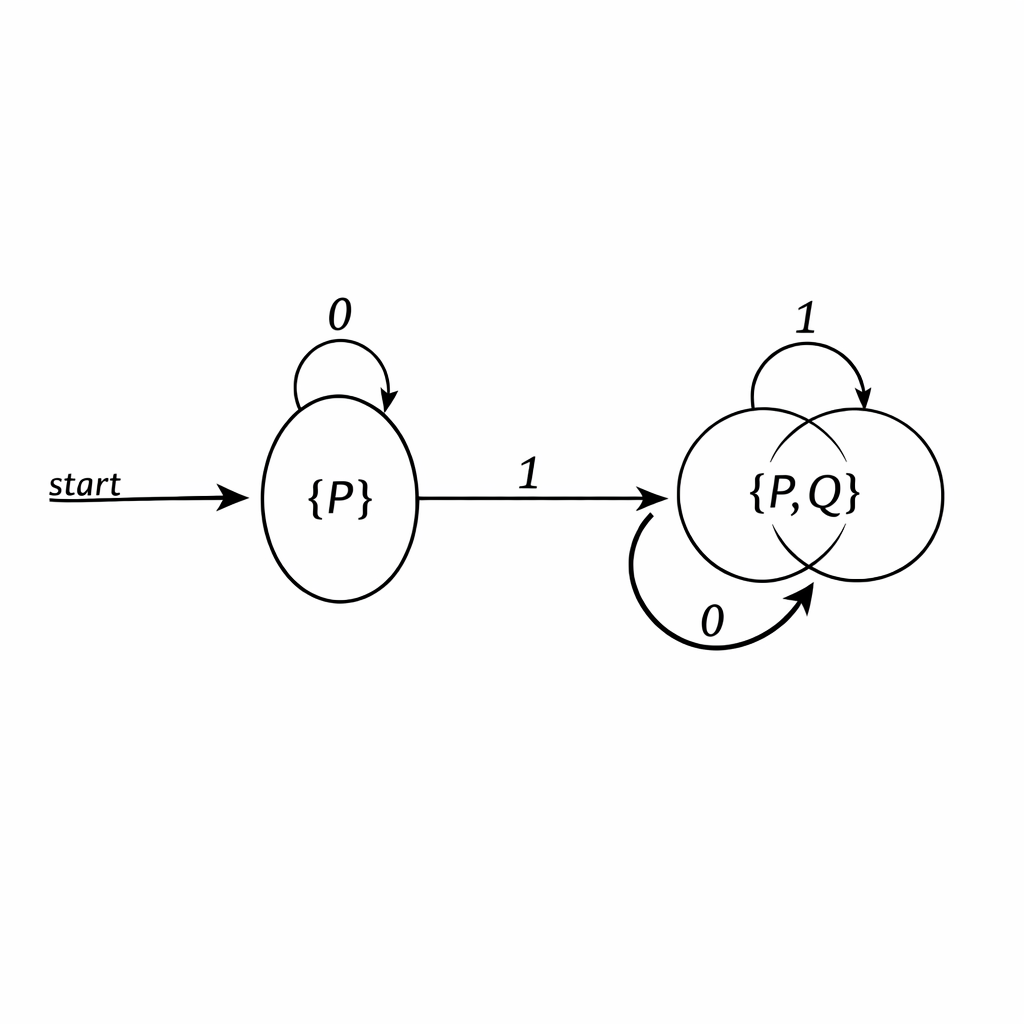}
\caption{Hand-drawn diagram (left), diagram from original description (middle) and diagram from edited description(right)}
\label{fig:dia3}
\end{figure}

For subjective evaluation, we present some samples of the original hand-drawn diagrams, diagrams generated from the original descriptions and diagrams generated from the edited descriptions in ~Figure \ref{fig:dia1}, and ~Figure \ref{fig:dia3}. We can observe that the diagrams generated from the edited descriptions tend to produce better results. 

\begin{figure}[h]
\centering
\includegraphics[width=0.3\textwidth]{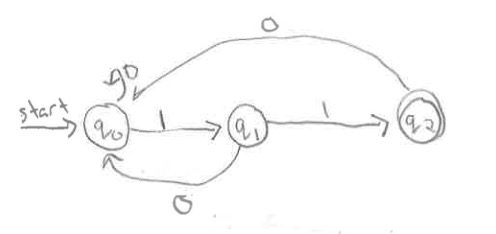}
\hfill
\includegraphics[width=0.3\textwidth]{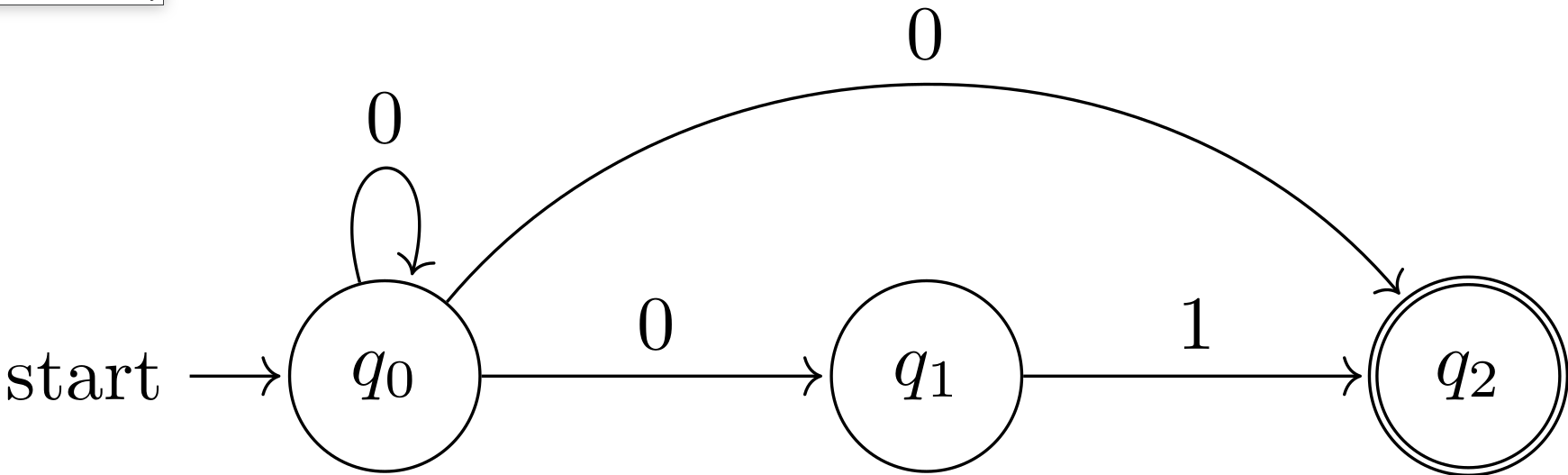}
\hfill
\includegraphics[width=0.3\textwidth]{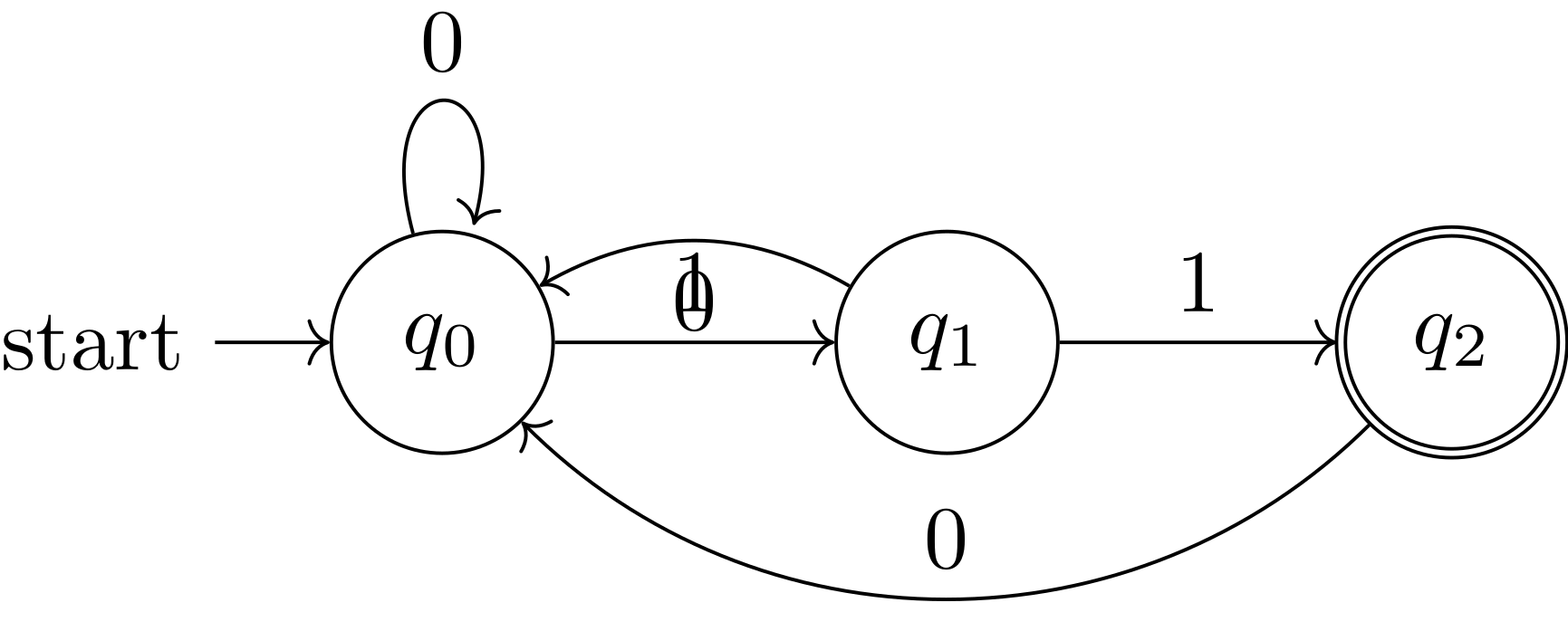}
\caption{Hand-drawn diagram (left), TikZ diagram from original description (middle) and TikZ diagram from edited description (right)}
\label{fig:dia4}
\end{figure}

For evaluation of images generated via TikZ, we present some samples of the original hand-drawn diagrams, TikZ compiled diagrams from the original descriptions and TikZ compiled diagrams from the edited descriptions in ~Figure \ref{fig:dia4}.

\section{Discussion}
The experimental results provide insight into the behavior of the reconstruction pipeline under different generation settings. Student-drawn diagrams exhibit variation in layout, shape, and notation. We found, that the vision-language model used often captures basic layout and labels, but frequently omit key structural components. Large language models can produce syntactically correct TikZ code, but correctness depends on the input text. Human editing has the potential to reduced such missing transitions, resolved label conflicts, and corrected start or accepting state errors. The editing process did not require advanced domain knowledge, but it did require careful visual inspection and step-by-step correction.

Our results reveal that the image via TikZ compilation produced better results compared to the direct image generation. The images created via TikZ compilation received an average score of 4.65 compared to the directly generated images which received an average score of 3.6.

\textbf{Prompt Engineering: }
Different prompt engineering strategies were adopted for accurate description generation using vision-language models. In particular, we compared model outputs prompted with and without access to the original exam question. Here we report our findings during the prompt optimization process. We found, descriptions generated with access to the exam question are more accurate. Question context reduced ambiguity about expected structure. We also report that one-shot prompts were more effective when the example and the target diagram shared similar layout or node configuration. Prompt format had a direct effect on the structure of the output.

\section{Conclusion}

This study evaluated a pipeline for reconstructing student drawn automata diagrams using vision language captioning, human revision, and TikZ based diagram generation. The results indicate that model generated descriptions before human revision frequently contain structural inaccuracies. Human correction substantially improves downstream reconstruction quality. TikZ based compilation achieved higher agreement with hand drawn diagrams than direct image generation from text.

In instructional settings, the proposed approach can support automated feedback by reconstructing a clean digital diagram from a student submission and identifying structural discrepancies relative to a reference solution. In assessment contexts, the method can assist grading by flagging missing transitions, incorrect accepting states, and other structural inconsistencies for instructor review.

This preliminary study is performed with limited amount of data and with a single vision-language model. As a future work we will curate more data, explore other models, examine automated detection of structural inconsistencies in model generated descriptions, refine prompt templates for different automata types, and perform large scale evaluation with formal agreement analysis.

\section*{Acknowledgment}

This research was partially supported by the NASA Established Program to Stimulate Competitive Research (EPSCoR), Grant No. 80NSSC22M0027 by the NASA West Virginia EPSCoR Committee.

\noindent Part of the computational requirements for this work were supported by resources at \textit{Delta at NCSA} and \textit{Anvil at Purdue University} through \textbf{ACCESS allocation Grant number: CIS250068 and CIS250766} from the \textit{Advanced Cyberinfrastructure Coordination Ecosystem: Services and Support (ACCESS)} program, which is supported by U.S. National Science Foundation grants \#2138259, \#2138286, \#2138307, \#2137603, and \#2138296.

\bibliographystyle{plain}
\bibliography{refs}

\end{document}